\documentclass[9pt,shortpaper,twoside,web]{ieeecolor}
\usepackage{generic}
\usepackage{cite}
\usepackage{amsmath,amssymb,amsfonts}
\usepackage{algorithmic}
\usepackage{graphicx}
\usepackage{textcomp}
\usepackage{tabularx}
\usepackage{mathrsfs}
\usepackage{booktabs}
\usepackage{makecell}
\usepackage[linesnumbered,ruled,vlined]{algorithm2e}

\def\BibTeX{{\rm B\kern-.05em{\sc i\kern-.025em b}\kern-.08em
    T\kern-.1667em\lower.7ex\hbox{E}\kern-.125emX}}
\begin{document}
\title{Accurate Scoliosis Vertebral Landmark Localization on X-ray Images via Shape-constrained Multi-stage Cascaded CNNs}
\author{Zhiwei Wang, Jinxin Lv, Yunqiao Yang, Yuanhuai Liang, Yi Lin, Qiang Li, Xin Li, and Xin Yang 
\thanks{Zhiwei Wang, Jinxin Lv,  Yuanhuai Liang, Qiang Li are with Britton Chance Center for Biomedical Photonics, Wuhan National Laboratory for Optoelectronics and with MoE Key Laboratory for Biomedical Photonics, Collaborative Innovation Center for Biomedical Engineering, School of Engineering Sciences, Huazhong University of Science and Technology, Wuhan, Hubei 430074, China. Zhiwei Wang and Jinxin Lv are the co-first authors contributing equally to this work. }
\thanks{Yunqiao Yang is with Department of Computer Science, City University of Hong Kong, Hong Kong, China.}
\thanks{Yi Lin is with Department of Computer Science and Engineering, The Hong Kong University of Science and Technology, Hong Kong, China.}
\thanks{Xin Li is with Department of Radiology, Union Hospital, Tongji Medical College, Huazhong University of Science and Technology, Wuhan, 430022, China.}
\thanks{Xin Yang is with School of Electronic Information and Communications, Huazhong University of Science and Technology, Wuhan, 430074, China.}
\thanks{Xin Li and Xin Yang are the corresponding authors (email:lxwsry2014@163.com, xinyang2014@hust.edu.cn).}}

\maketitle

\begin{abstract}
Vertebral landmark localization is a crucial step for variant spine-related clinical applications, which requires detecting the corner points of 17 vertebrae. However, the neighbor landmarks often disturb each other for the homogeneous appearance of vertebrae, which makes vertebral landmark localization extremely difficult.
In this paper, we propose multi-stage cascaded convolutional neural networks (CNNs) to split the single task into two sequential steps, i.e., center point localization to roughly locate 17 center points of vertebrae, and corner point localization to find 4 corner points for each vertebra without distracted by others.
Landmarks in each step are located gradually from a set of initialized points by regressing offsets via cascaded CNNs. Principal Component Analysis (PCA) is employed to preserve a shape constraint in offset regression to resist the mutual attraction of vertebrae. We evaluate our method on the AASCE dataset that consists of 609 tight spinal anterior-posterior X-ray images and each image contains 17 vertebrae composed of the thoracic and lumbar spine for spinal shape characterization. Experimental results demonstrate our superior performance of vertebral landmark localization over other state-of-the-arts with the relative error decreasing from $3.2e{-3}$ to $7.2e{-4}$.
\end{abstract}

\begin{IEEEkeywords}
vertebral landmarks, multi-stage, cascaded, shape constraint
\end{IEEEkeywords}

\section{Introduction}
\label{sec:introduction}
Vertebral landmark localization is a key step for computer-aided diagnosis of spinal diseases,~e.g., Cobb angle calculation, bio-mechanical load analysis, detecting vertebral fractures, or other pathologies~\cite{bayat2019vertebral}.
However, manually identifying 17 vertebrae and localizing 4 corner points of each vertebra are extremely time-consuming and infeasible for large-scale screening.
Thus, the automatic methods that detect vertebral landmarks, i.e., $17 \times 4 = 68$ points in total, are in high demand. 
Although automatic vertebral landmark localization has been studied for decades, it still remains challenging due to the high ambiguity and variability of X-ray image~\cite{sun2017direct}, the superimposing of heterogeneous soft tissue, and the similar texture feature of nearby vertebrae~\cite{bayat2019vertebral}.

To address this challenging task, there exist two major families of approaches, i.e., the heatmap-based approach and the regression-based approach.

{\bf Heatmap-based approach} is one of the most successful approaches of vertebral landmark localization.
It typically utilizes a convolutional neural network (CNN) to generate a heatmap with the same size of the input posterior-anterior X-ray image.
Those vertebral landmarks are then obtained by finding points with a local maximum response in the heatmap. 
Zhang~\emph{et al.}~\cite{zhang2021automated} utilized four different fully-connected layers to predict four heatmaps, each of which indicates a group of corner points with different semantics, i.e., top left, top right, bottom left, and bottom right points. 
The coordinates of landmarks are obtained indirectly by post-processing the predicted heatmap, i.e., finding local maximums' column and row indexes as landmark's coordinates.
Those indexes are discrete integers, leading to a non-trivial quantization errors to the continuously numerical ground-truth~\cite{nibali2018numerical}.
To address this problem, Yi~\emph{et al.}~~\cite{yi2020vertebra} proposed to first locate 17 center points of the whole spine based on heatmaps, and then regress numerical coordinates of corner points relative to those center points using shape regression.

Despite the success of these heatmap-based methods, a common flaw can be often observed from their results, i.e., scattering false positives incurred by other vertebra-like structures and missing landmarks induced by dilution of superimposed tissues.
These incorrect predictions derive wrong clinical parameters, resulting in misdiagnosis.
On the other hand, heatmap-based landmark localization approaches have been demonstrated to be robust in tasks with large pose variations,~e.g., human pose estimation, However, they could be overkilled in our task since human spines roughly fit a similar curve, that is, the pose variation is secondarily important.

\begin{figure*}[!t]
	\centering
	\includegraphics[width=1.0\textwidth]{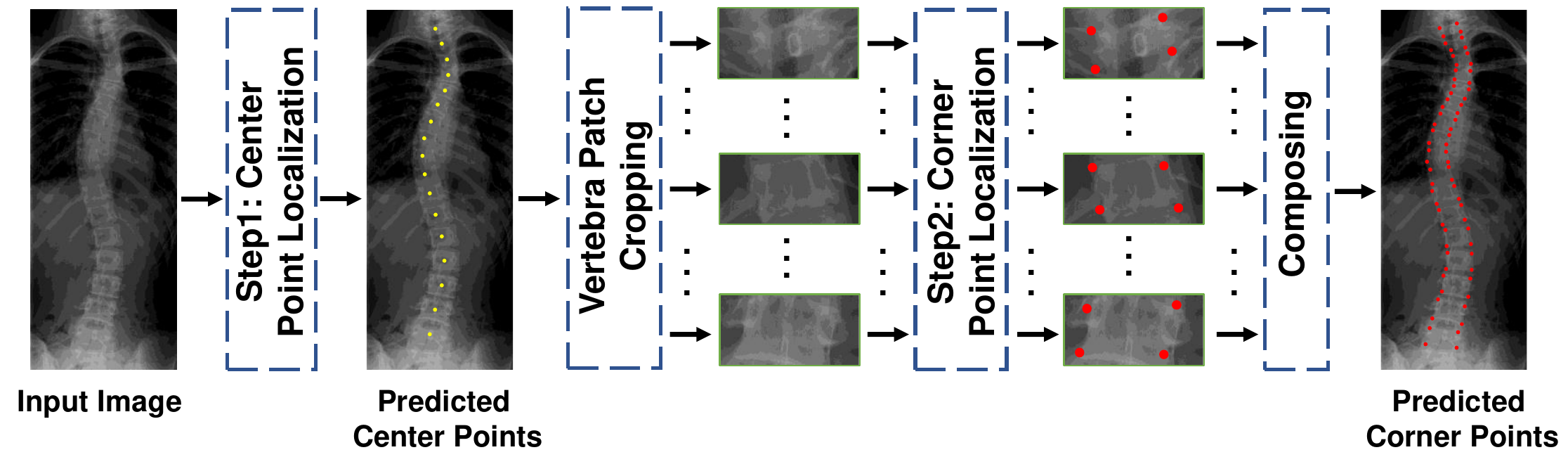}
	\caption{Overview of the proposed two-step vertebral landmark localization methods. See Fig.~\ref{step1} and Fig.~\ref{step2} for details of the two steps.}
	\label{fig_framwork}
\end{figure*}

For the task of landmark localization, there also exist several works resorted to the Transformer technique, which was proposed originally in the field of nature language processing (NLP) and are flourishing recently in a variety of computer vision fields including landmark localization.
Most, if not all, transformer-based methods typically follow the basic pipeline of the heatmap-based approach, and also generate heatmaps to obtain landmarks' coordinates.
Yang~\emph{et al.}~\cite{yang2021transpose} incorporated a transformer in to a CNN to implicitly capture long-range spatial relationships of human body parts efficiently, and then decoded those relationships into heatmaps for human pose estimation.
Zhao~\cite{zhao2021graformer} focused on 3D landmark localization, and combined the strengths of graph convolution network (GCN) and transformer to characterize a set of pre-localized 2D landmarks so as to map them into their 3D counterparts.
However, these transformer-based methods involves a self-attention mechanism, and have to compute similarity scores of all pairs of landmarks, which is prohibitively computationally expensive and meanwhile requires a large amount of data for training.
To address this, Tao~\cite{tao2021spine} proposed a light-weight transformer for vertebra detection in 3D CT volumes, but vertebral landmark localization in 2D X-ray images is more challenging than that in 3D CT volumes due to the superimposing tissues and less spatial information.
The effectiveness of the transformer on vertebral landmark localization has not been studied and verified.

{\bf Regression-based approach} is another family of vertebral landmark localization, which utilizes an end-to-end CNN model to directly regress the coordinates. 
Sun~\emph{et al.}~\cite{sun2017direct} proposed structured support vector regression ($\mathrm{S^2VR}$) which simultaneously encourages multiple outputs to share similar sparsity patterns such as spatial correlations. 
Wu~\emph{et al.}~\cite{wu2017automatic} constructed a relation matrix for spinal landmarks to explicitly enforce the dependencies between outputs. 
Although these methods can ensure one-to-one matches between the regressed points and ground truth points, directly regressing coordinates has two major problems: (1) CNNs may not be sufficiently powerful to precisely regress all landmarks in a single feed-forward, and (2) points of each vertebra tend to be strongly attracted by other vertebrae during regression since they all share similar appearances. 
The affinity among vertebrae could cause a sudden point shifting to its neighbors, which can hardly be corrected if the shifted points are too far from their ground-truth.

In this paper, we propose multi-stage cascaded CNNs for vertebral landmark localization, which consists of two sequential steps,  i.e., {\em center point localization} for vertebrae and {\em corner point localization} for each localized vertebra as shown in Fig.~\ref{fig_framwork}. 
In the first step, we train three-stage cascaded CNNs to gradually update 17 center points of a vertebrae from 17 initial starting points. The output points of each former stage CNN is used as input of the latter one. 
In each stage of center point localization, local features around each vertebra are first extracted separately and then concatenated to jointly regress all center points of the vertebrae. 
After the first step, each vertebra is roughly localized to guide the following corner point localization step focus on its own area, overcoming the point shifting problem caused by the affinity among vertebrae. 
In the second step, another three-stage cascaded CNNs are trained to gradually localize 4 corner points for each vertebra, and in turn output the final 68 landmarks. 
To further address the mutual attraction problem among vertebrae during regression, we propose to regress the PCA-transformed coordinate offsets instead of the direct absolute ones using a CNN in each stage. In this way, the location of a single point can be constrained using a global shape constraint,~e.g., a pre-defined variance of distance between two adjacent vertebrae.

\begin{figure*}[!t]
	\centering
	\includegraphics[width=0.8\textwidth]{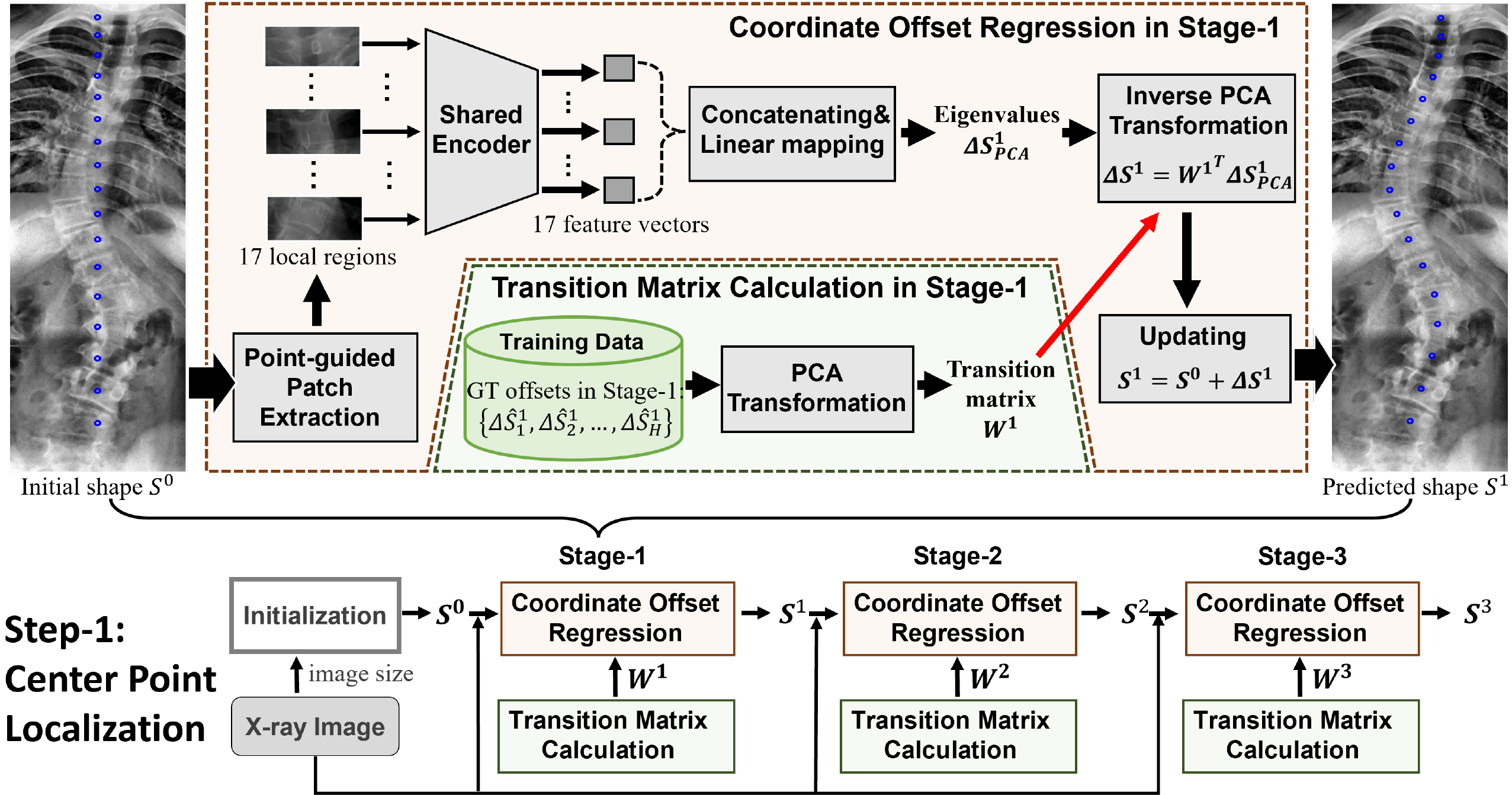}
	\caption{Overview of the multi-stage cascaded framework in the first step, i.e., center point localization. 
	}
	\label{step1}
\end{figure*}

To summarize, our key contributions are as follows:
\begin{itemize}
	\item [1)] 
	We propose a multi-stage regression framework for accurate and robust vertebra landmark localization, where the former stages identify each vertebra to keep the latter stages from being disturbed by neighboring vertebrae.       
	\item [2)]
	We regress PCA-transformed offsets using CNNs in each stage to progressively guide initially regressed points to the true target positions under a global shape constraint. To the best of our knowledge, it is the first try to utilize multi-stage cascaded CNNs for shape-constrained vertebral landmark localization instead of resorting to those overused heatmap-based approaches.
	\item [3)]
	Extensive experimental results on the public dataset demonstrate that our method enjoys a benefit of much fewer false positives as well as missing landmarks, and thus achieves superior performance to other state-of-the-arts~\cite{Simonyan2015VeryDC,sun2017direct,wu2017automatic,yi2020vertebra,zhang2021automated} by decreasing the relative error from $3.2e{-3}$ to $7.2e{-4}$.
\end{itemize}

\section{Method}
As shown in Fig.~\ref{fig_framwork}, we address the task of 68 vertebral landmark localization by two sequential steps, i.e., 17 center point localization for the vertebral column and 4 corner point localization for each vertebra.
In both steps, we propose a three-stage cascaded framework to gradually update an initial shape (i.e., points) to the ground-truth (GT) shape.
Each stage contains two major modules, i.e., \emph{Transition Matrix Calculation} to mine shape constraints from the training set by PCA, and \emph{Coordinate Offset Regression} to update the shape from the previous stage (initial shape if in the first stage).

In the following, we detail the two steps and the two modules respectively.

\subsection{Step-1: Three-stage Framework of Center Point Localization}

Our goal in this step is to gradually localize 17 center points on the whole X-ray image from a set of starting points, i.e., $S^0 = (x^0_1,y^0_1,...,x^0_{17},y^0_{17})$. 
To this end, the input image is first proportionally resized to 680-pixel height.
The initial shape $S^0$ is placed based on the image size, and then gradually updated via three stages.
In each stage (as shown in Fig.~\ref{step1}), Transition Matrix Calculation first mines a transition matrix from the GT offsets to preserve the shape constraint in regression.
Coordinate Offset Regression regresses several eigenvalues, which are then converted to the shape offset by the transition matrix. 
Updated shape is obtained by applying the shape offset on the predicted shape from the previous stage (initial shape if in the first stage).
Three stages are cascaded to gradually update $S^0$ as $S^{n}=S^{n-1}+ \Delta S^{n}$, where $n=1,2,3$, as shown in the bottom of Fig.~\ref{step1}.


\subsubsection{Transition Matrix Calculation}
The target of prediction in this step is 34 coordinates of the 17 center points, while most of them are redundant since the vertebral column roughly fits a curve. 
In landmark localization, the redundancy in the regression target could greatly increase, the chance of losing a shape constraint due to the uncontrollable regression process of each landmark. 
	PCA~\cite{abdi2010principal,jolliffe2016principal}is a widely-used dimensional reduction approach that can be applied to capture the underlying shape constraint by the first few principal components (i.e., eigenvalues). 
In addition, the PCA-mined shape constraint also can tolerate noises in landmarks since landmark misalignments usually contribute little to the first few principal components. 
In this work, we apply PCA to reduce the 34-dimensional regression target to a few eigenvalues, which is detailed as follows.

Given the input shape $S^{n-1}$ in the $n$-th stage (initial shape $S^0$ if in the first stage), the original regression target $\Delta \hat{S}^n$ is the coordinate offset between $S^{n-1}$ and the GT shape $\hat{S}$, which is calculated as:
\begin{equation}
	\Delta \hat{S}^n=\hat{S}-S^{n-1} \in \mathbb{R}^P, n=1,2,3
	\label{eq:1}
\end{equation}
By collecting $\Delta \hat{S}^n$ of all $H$ training images, we can get a matrix $\Delta \hat{\mathbb{S}}^n$ with the size of $P\times H$, where each column represents a training sample, and each row indexes a coordinate, that is, $P=34$.
The covariance matrix  $\mathbf{C}$ is then calculated in Eq.~(\ref{eq_cov}):
\begin{equation}
	\mathbf{C} = \frac{1}{P} \Delta \hat{\mathbb{S}}^n(\Delta \hat{\mathbb{S}}^n)^{\mathrm{T}} \in \mathbb{R}^{P \times P}
	\label{eq_cov}
\end{equation}
Next, we calculate the eigenvalues and their corresponding eigenvectors of the covariance matrix $\mathbf{C}$ which satisfy the  Eq.~(\ref{eq_eigen}):
\begin{equation}
	\lambda_j \mathbf{q}_j = \mathbf{C} \mathbf{q}_j, j=1,...,P
	\label{eq_eigen}
\end{equation}
where $\lambda_j$ denotes the $j$-th eigenvalue and we have $\lambda_1 > \lambda_2 > ... > \lambda_P$, $\mathbf{q}_j$ denotes the corresponding eigenvector.
Accordingly, the PCA-derived trainsition matrix $W^{n}$ can be constructed by selecting the first $Q$ eigenvetors:
\begin{equation}
	W^{n}=[\mathbf{q}^T_1;\mathbf{q}^T_2;...;\mathbf{q}^T_Q] \in \mathbb{R}^{Q \times P}, Q \ll P, n=1,2,3
	\label{eq:4}
\end{equation}

The calculated transition matrix $W^{n}$ in the $n$-th stage can be used to transform the original regression target $\Delta \hat{S}^n$ into $Q$ eigenvalues according to Eq.~(\ref{transformation}):
\begin{equation}
	\Delta \hat{S}^n_{PCA} = W^n \Delta \hat{S}^n \in \mathbb{R}^Q, n=1,2,3
	\label{transformation}
\end{equation}

Comparing Eq.~(\ref{transformation}) and Eq.~(\ref{eq:1}), we can find that PCA allows us to regress $Q$ eigenvalues instead of $P$ coordinates ($Q=8$ and $P=34$ in the first step).
Each eigenvalue controls the change of a shape in the corresponding eigenvector-defined direction, e.g., bending, point density, rotation,~etc.
In this way, we can avoid dramatic differences between the estimated center points in each stage and those in training images.

\subsubsection{Initialization}
After calculation of the transition matrix, we perform Coordinate Offset Regression to predict a shape offset for updating.
In each stage, the prediction of shape offset conditions on the input shape predicted in the previous stage.
Therefore, the shape has to be initialized before the \emph{first} stage.

Based on all shapes in the training set, we calculate a normalized averaged shape $\overline{S}^{norm}$ for initialization.
To this end, each GT shape is normalized according to the size of the corresponding X-ray image:
\begin{equation}
	\hat{S}_{h}^{norm}=\text{normalize}(\hat{S}_{h})=(\frac{x_1}{\text{Wid}},\frac{y_1}{\text{Hei}},...,\frac{x_{17}}{\text{Wid}},\frac{y_{17}}{\text{Hei}})_h, h=1,...,H
\end{equation}
where $h$ indexes the GT shape in the training data, $H$ is the total number of training images, $\hat{S}_{h}$ is the unnormalized GT shape, $(x_i,y_i)$ is the coordinates of the $i$-th landmark, Wid and Hei indicates the width and the height of the $h$-th X-ray image. 

After normalization, every coordinate is converted to that ranging from 0 to 1. Then, we calculate the normalized averaged shape as:
\begin{equation}
	\bar{S}^{norm} = \frac{1}{H}\sum_{h=1}^{H} \hat{S}^{norm}_h
	\label{eq:7}
\end{equation}

To initialize in the first stage, we de-normalize $\bar{S}^{norm}$ according to the size of input X-ray image as:
\begin{equation}
	\begin{aligned}
		S^0 &= \text{de-normalize}(\bar{S}^{norm})\\
		&=(\bar{x}^{norm}_1 \times \text{Wid},\bar{y}^{norm}_1 \times \text{Hei},...,\bar{x}^{norm}_{17} \times \text{Wid},\bar{y}^{norm}_{17} \times \text{Hei})
	\end{aligned}
	\label{eq:8}
\end{equation}
where $(\bar{x}^{norm}_i,\bar{y}^{norm}_i )$ is the normalized coordinates of the $i$-th landmark of $\bar{S}^{norm}$.
Note that Wid and Hei could vary with different input images, the initial shape thus changes correspondingly. 

\subsubsection{Coordinate Offset Regression}
Coordinate Offset Regression sequentially performs feature extraction, regression, and updating based on the input shape $S^{n-1}$ as shown in Fig.~\ref{step1}.

{\bf Regression:}  In each stage, we extract only features in a local region at every center point from the predicted or initialized shape of the previous stage to capture highly discriminative and noise-resistable features.

\begin{table*}[!t]
	\begin{footnotesize}
		\caption{The architecture of the truncated MobileNet-V2 for feature extraction in both steps. The size of cropped local regions is $48 \times 80$ and $48 \times 48$ in the two steps respectively. $t$: the channel increasing factor in InvertedResidualBlock, $c$: output channel number, $n$: the number of the operator repeating, $s$: the stride, Global average pooling (GAP) spatially squeezes the feature maps.}
		\label{architecture}
		\begin{center}
			\renewcommand\arraystretch{1.2}
			\setlength{\tabcolsep}{2mm}{
				\begin{tabularx}{0.9\linewidth}{p{2.5cm}|p{2.5cm}|p{3.5cm}|*4{>{\centering\arraybackslash}X}}
					\toprule
					Input size in step-1 & Input size in step-2 &  Operator & $t$ & $c$ & $n$ & $s$ \\ \midrule
					$48 \times 80 \times 1$ & $48 \times 48 \times 1$ & Conv2d $1 \times 1$ & - & 32 & 1 & 2 \\
					$24 \times 40 \times 32$ & $24 \times 24 \times 32$ & InvertedResidualBlock & 1 & 16 & 1 & 1 \\
					$24 \times 40 \times 16$ & $24 \times 24 \times 16$ & InvertedResidualBlock & 6 & 24 & 2 & 2 \\
					$12 \times 20 \times 24$ & $12 \times 12 \times 24$ & InvertedResidualBlock & 6 & 32 & 3 & 2 \\
					$6 \times 10 \times 32$ & $6 \times 6 \times 32$ & InvertedResidualBlock & 6 & 64 & 4 & 2 \\
					$3 \times 5 \times 64$ & $3 \times 3 \times 64$ & InvertedResidualBlock & 6 & 96 & 3 & 1 \\
					$3 \times 5 \times 96$ & $3 \times 3 \times 96$ & InvertedResidualBlock & 6 & 160 & 3 & 2 \\
					$2 \times 3 \times 160$ & $2 \times 2 \times 160$ & InvertedResidualBlock & 6 & 320 & 1 & 1 \\
					$2 \times 3 \times 320$ & $2 \times 2 \times 320$ & Conv2d $1\times1$ & - & 64 & 1 & 1 \\
					$2 \times 3 \times 64$ & $2 \times 2 \times 64$ & GAP & - & - & 1 & - \\ \midrule
					Output: 64 & Output: 64 & - & - & - & - & - \\ \bottomrule
				\end{tabularx}
			}
		\end{center}
	\end{footnotesize}
\end{table*}

In details, we crop 17 local regions size of $80 \times 192$ guided by those predicted or initialized center points and then resize them to $48 \times 80$ for feature extraction by the shared encoder, i.e., a truncated MobileNet-V2~\cite{sandler2018mobilenetv2}.
Table~\ref{architecture} gives the architecture details of our employed truncated MobileNet-V2.
The Inverted Residual Block shown in Table~\ref{architecture} first increases the number of channels by $t$ times, and then convolves the feature maps by a depth-wise convolutional layer with the stride of $s$ (parameter $s$ only works at the first time, and $s=1$ in the following $n-1$ repeats), and at last decreases the number of channels to $c$.
Each InvertedResidualBlock operation is repeated by $n$ times.
More details of the Inverted Residual Block can refer to~\cite{sandler2018mobilenetv2}. 

As shown in the first column of Table ~\ref{architecture}, the shared encoder extracts a 64-dimensional feature vector for each local region, which totally yields 17 feature vectors in the first step.
Since both local information of each landmark and shape context of other landmarks matter~\cite{ren2014face}, we then concatenate these local feature vectors to form a 1088-d feature vector, followed by a fully connected layer to linearly regress a few eigenvalues $\Delta S^{n}_{PCA}$ as:
\begin{equation}
	\Delta S^{n}_{PCA} = R(S^{n-1};\theta^n)
	\label{eq:10}
\end{equation}
where $\theta^n$ is the parameters of the shared encoder and the fully connected layer in the $n$-th stage, and $R(S^{n-1};\theta^n)$ means regression conditioning on the input shape $S^{n-1}$.


{\bf Updating:} The numerical coordinates offsets for the $n$-th stage stage, i.e., $\Delta S^n = (\Delta x^n_1,\Delta y^n_1,...,\Delta x^n_{17},\Delta y^n_{17})$, are obtained by the inverse PCA  transition matrix ${W^n}^T$ as:
\begin{equation}
	\Delta S^n = {W^n}^T\Delta {S}^n_{PCA}
	\label{eq:10-1}
\end{equation}
where $W^n$ is calculated beforehand as in Transition Matrix Calculation, and $T$ denotes the transposition operation. 
The coordinates of those center points from the previous stage $S^{n-1}$ are thus updated as:
\begin{equation}
	S^n = S^{n-1}+\Delta S^n
	\label{eq:10}
\end{equation}

\begin{algorithm}[!t]
	\begin{footnotesize}
		\label{alg:2}
		\caption{Pseudo-code for the testing phase of Step-1.}
		\KwIn{A X-ray image with the size of Hei $\times$ Wid.}
		\KwOut{Final prediction $S^3$.}	
		
		\textbf{Initialization:}
		Get $S^0$ as in Eq.~(\ref{eq:8});
		
		\For{n=1,2,3}{
			\textbf{Regression}: $\Delta S^{n}_{PCA} = R(S^{n-1};\theta^n)$;
			
			$\Delta S^n = {W^n}^T\Delta {S}^n_{PCA}$;
			
			\textbf{Updating:} $S^n = S^{n-1}+\Delta S^n$
		}
		
		Return $S^3$;
	\end{footnotesize}
\end{algorithm}

The testing phase is summarized in Algorithm~\ref{alg:2}.
After obtaining $S^n$ in the $n$-th stage, we go back to the regression process and crop a new set of local regions guided by $S^n$ in the $(n+1)$-th stage.
Since $S^n$ is closer to the GT shape than $S^{n-1}$, the features extracted are more discriminative accordingly, yielding a more accurate shape offset $\Delta S^{n+1}$ for updating in the next stage.
With such stage-by-stage strategy, the shape is updated from $S^0$ to $S^3$, closer to the GT shape gradually.

We optimize $\theta^n$ in each stage in the training phase as described below.

{\bf Optimization:} 
The training phase of the three-stage cascaded CNN for center point localization is summarized in Algorithm~\ref{alg:1}.
In the $n$-th stage, we first obtain the GT eigenvalues $\Delta \hat{S}^n_{PCA}$ by Eq.~(\ref{transformation}), and then predict eigenvalues $\Delta {S}^n_{PCA}$ based on the previously predicted or initialized shape $S^{n-1}$.
The smooth L1 loss function~\cite{girshick2015fast} is employed to calculate the loss $\mathcal{L}$ between the GT $\mathrm{\Delta}{\hat{S}}_{PCA}^n$ and the predicted $\mathrm{\Delta}S_{PCA}^n$ as:
\begin{equation}
	\begin{array}{lr}
		\mathcal{L}=\sum_{i}{{\rm Smooth}_{L_1}({\mathrm{\Delta}S_{PCA}^n}_i-{\mathrm{\Delta}{\hat{S}}_{PCA}^n}_i)} \\
		\\
		{\rm Smooth}_{L_1}(x)=
		\left\{
		\begin{array}{lr}
			\frac{0.5x^2}{\beta}, &\rm{if}~|x| < \beta  \\
			|x|-0.5\beta, &\rm{otherwise}		 
		\end{array}
		\right.
	\end{array}
	\label{eq:12}
\end{equation}
where $i$ indexes each eigenvalue, and $\beta=0.001$ is used to prevent gradient explosions.
Parameters $\theta^n$ in both truncated MobileNet-V2 and fully connected layer are optimized by minimizing $\mathcal{L}$.

After training, we have three transition matrices $W^1$, $W^2$, $W^3$ and the learned parameters $\theta^1$, $\theta^2$, $\theta^3$ for three cascaded stages, as well as a normalized averaged shape $\overline{S}^{norm}$ for initialization.

\begin{algorithm}[!t]
	\begin{footnotesize}
		\label{alg:1}
		\caption{Pseudo-code for the training phase of Step-1.}
		\KwIn{All training images and their GT center points.}
		\KwOut{Transition Matrix $W^1$, $W^2$,$W^3$ and CNNs $R(;\theta^1)$,$R(;\theta^2)$,$R(;\theta^3)$ for three cascaded stages; A normalized averaged shape $\overline{S}^{norm}$.}	
		
		Calculation the normalized averaged shape $\overline{S}^{norm}$ as in Eq.~(\ref{eq:7})
		
		\textbf{Initialization:}
		Get $S^0$ for every training sample as in Eq.~(\ref{eq:8});
		
		\For{n=1,2,3}{
			Calculation of the transition matrix $W^n$ as in Eq.~(\ref{eq:4});
			
			\For{each training sample in each epoch}{
				
				Calculation of PCA-transformed regression target $\Delta \hat{S}^n_{PCA}$ as in Eq.~(\ref{transformation});
				
				\textbf{Regression}: $\Delta S^{n}_{PCA} = R(S^{n-1};\theta^n)$;
				
				\textbf{Optimization:}			
				Calculation of $\mathcal{L}$ between $\Delta S^{n}_{PCA}$ and $\Delta \hat{S}^n_{PCA}$ as in Eq.~(\ref{eq:12}) to optimize $\theta^n$;
			}
			
			Get $\Delta S^n$ for every training sample as in Eq.~(\ref{eq:10-1});
			
			\textbf{Updating:} $S^n = S^{n-1}+\Delta S^n$ for every training sample;
		}
		
		Return $\{ W^1, W^2, W^3\}$, $ \{ R(;\theta^1), R(;\theta^2), R(;\theta^3) \}$, and $\overline{S}^{norm}$;
	\end{footnotesize}
\end{algorithm}

\subsection{Step-2: Three-stage Framework of Corner Point Localization}

\begin{figure*}[!t]
	\centering
	\includegraphics[width=0.85\textwidth]{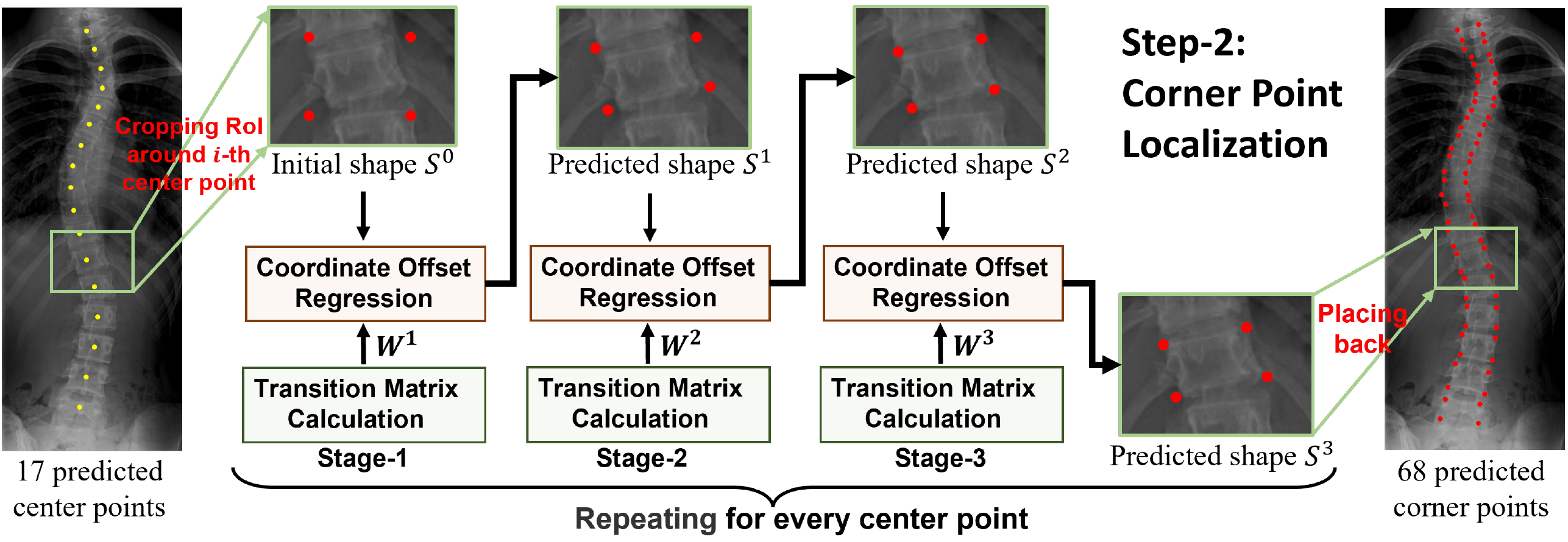}
	\caption{Overview of the multi-stage cascaded framework in the second step, i.e., corner point localization. 17 RoIs are cropped from the X-ray image at the predicted center points. For each RoI, the localization process is mostly the same as the first step (see Fig.~\ref{step1}) with only two differences, (1) the number of points is 4 instead of 17, (2) initialization is based on the size of RoI instead of the size of X-ray image. By repeating the process for every RoI, 68 corner points can be obtained eventually.}
	\label{step2}
\end{figure*}

After localization of 17 center points in the first step, we then localize 4 corner points for each vertebra in the second step.
Fig.~\ref{step2} visualizes the overview of the three-stage cascaded framework for corner points localization.
As can be seen, most parts are same as the first step except that we first extract 17 regions of interest (RoIs) at every predicted center point and then repeatedly perform 4 corner points localization on each RoI instead of the whole X-ray image.

Specifically, a RoI size of $80 \times 96$ is extracted with its center location aligned with one predicted center point.
We perform Initialization, Regression and Updating on the extracted RoI as explained in the Step.
Four initial points are placed based on the size of RoI instead of the whole X-ray image.
Four local regions size of $48 \times 48$ for feature extraction in Regression are cropped at the predicted or initialized corner points.
After prediction of 4 corner points for each vertebra, we place them on the X-ray image and finally obtain 68 corner points as shown in Fig.~\ref{step2}.

\subsection{Training Details}
We use adaptive CLAHE~\cite{reza2004realization} to enhance the X-ray images and apply horizontally flipping to augment the amount of training data by twice.
We utilize the smooth $L_1$ loss with $\beta= 0.001$, and the Adam optimizer with an initial learning rate equal to $5e{-4}$ for training.
We decay the learning rate by $0.5$ after each epoch.
We separately train both step in a maximum of 5 stages and 8 epoches for each stage.
Our framework is implemented in PyTorch and trained and tested on a computational platform with a TITAN XP GPU.

\section{Experiments and Results}
\subsection{Dataset and Evaluation Metric}
We use the public AASCE dataset\footnote{https://aasce19.github.io/\label{web}} that consists of 609 tight spinal anterior-posterior X-ray images and each image contains 17 vertebrae composed of the thoracic and lumbar spine for spinal shape characterization. Each vertebra is identified by its four corners, resulting in 68 points per spinal image. These landmarks were manually annotated based on visual cues. 
We utilize two different training-test splits for evaluation, i.e., (1) Official Split\textsuperscript{\ref{web}}, where the dataset is split into 481 images for training and the rest 128 images for validation, as well as additional 98 challenge-private images with no annotations for test, (2) Consistent Split~\cite{wu2017automatic}, where only the official training set is used and split into 431 for training and 50 for test, which is consistent with existing state-of-the-arts ~\cite{Simonyan2015VeryDC,sun2017direct,wu2017automatic,zhang2021automated}.

To compare with previous works for vertebral landmark localization, we utilize the normalized mean squared error (MSE) as the evaluation metric, which is calculated as
\begin{equation}
	f_{mse}=\frac{1}{68}\sum_{i=1}^{68}(\hat{s}_i -s_i)^2  |  s_i \in S^{norm}; \hat{s}_i \in \hat{S}^{norm}
\end{equation}
where  $i$  is the index of each value in $S^{norm}$ and $\hat{S}^{norm}$, $S^{norm}$ and $\hat{S}^{norm}$ are normalized estimated coordinates and normalized GT coordinates, respectively.

In addition, the AASCE challenge was initialized for the purpose of estimation of three Cobb angles, i.e., proximal thoracic (PT) angle, main thoracic (MT) angle and the thoracolumbar (TL) angle~\cite{wang2021evaluation}, which is a downstream application of vertebral landmark localization.
Therefore, we also evaluate our method on the task of Cobb angle estimation.
The evaluation metric is Symmetric Mean Absolute Percentage (SMAPE), which is calculated as:
\begin{equation}
	SMAPE=\frac{1}{N}\sum_{j}^{N}\frac{\sum_{i=1}^{3}{(\left|a_{ji}-b_{ji}\right|)}}{\sum_{i=1}^{3}{(a_{ji}+b_{ji})}}
\end{equation}
where $i$ indexes the three Cobb angles, $j$ denotes the $j$-th image, and $N$ is the total number testing images.
$a$ and $b$ are the predicted and the ground-truth Cobb angles respectively.

\subsection{Results}
\subsubsection{Comparison with state-of-the-arts for vertebral landmark localization}

We evaluate our method on AASCE with the Consistent Split,  which is consistent with previous vertebral landmark localization works.

\begin{figure*}[!t]
	\centering
	\includegraphics[width=0.8\textwidth]{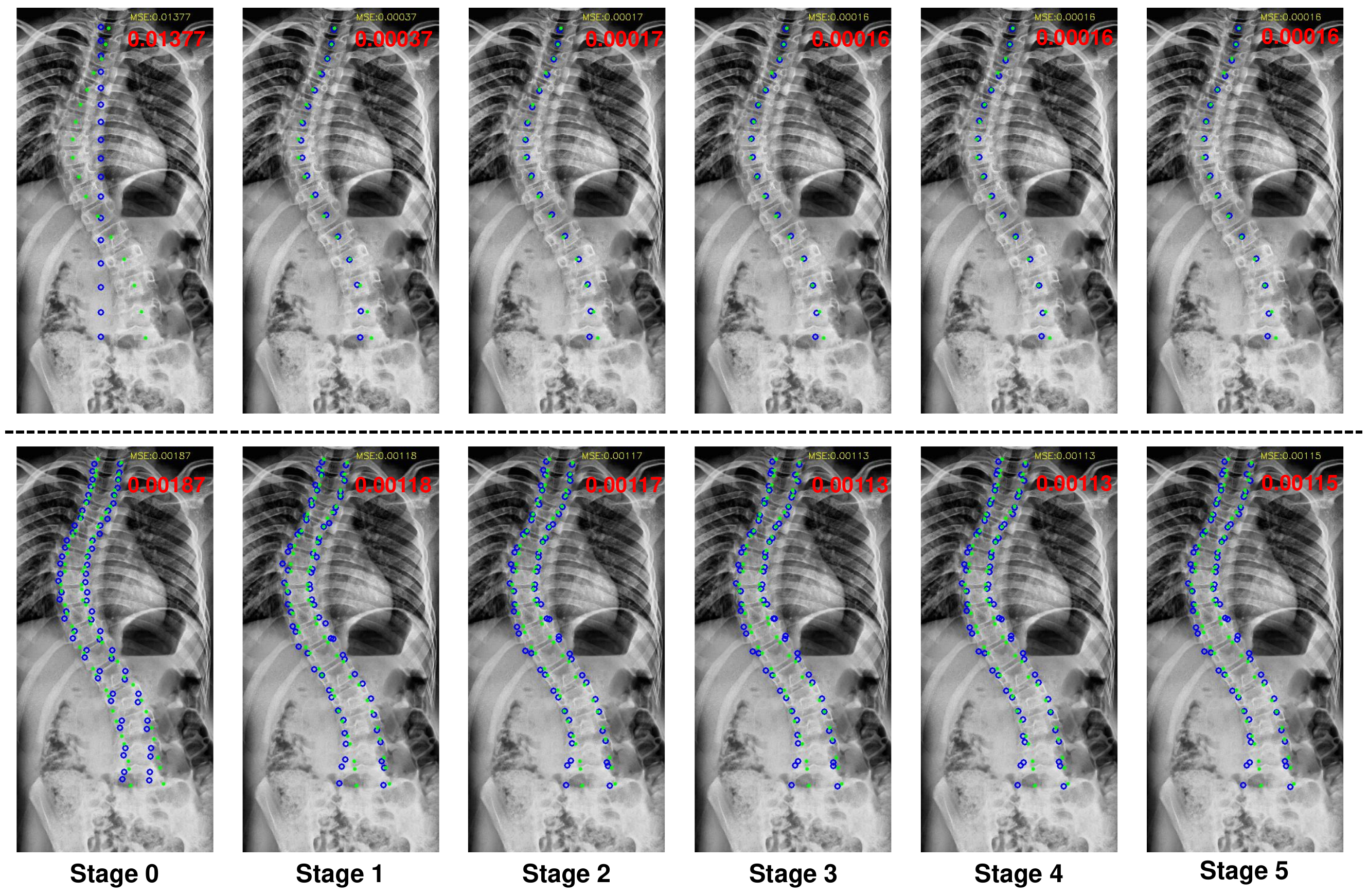}
	\caption{The top row and bottom rows show the localization results of center points and corner points, respectively. The green dots and blue circles indicate ground truth and prediction, respectively. The red numbers indicate the MSE.}
	\label{Visualization}
\end{figure*}

{{\bf Visualization.}}
We show the visual results of landmark localization in Fig.~\ref{Visualization}.
It can be seen that for both center points localization and corner points localization, our three-stage cascaded framework can gradually locate almost exactly the same points as the GT corners of vertebrae stage-by-stage, which well recover both global and detailed structures of a spine. 
Besides, for both steps, the effect of the first several stages is the most obvious, while the results maintain stability after the third stage, thus we take the prediction of the third stage of each step as the final result of our method.

{{\bf Comparison.}}
We compare our proposed framework with several state-of-the-art methods~\cite{Simonyan2015VeryDC,sun2017direct,wu2017automatic,yi2020vertebra,zhang2021automated}. 
Among them, only Yi~\emph{el al.}~\cite{yi2020vertebra} have released their source code, therefore, we used the source code for training and testing based on Consistent Split.
The results of $\mathrm{S^2VR}$~\cite{sun2017direct}, Conv with dense~\cite{Simonyan2015VeryDC}, BoostNet~\cite{wu2017automatic} and SLSN~\cite{zhang2021automated} are borrowed from the paper of SLSN, and we all followed the same standard of data split.

\begin{table}[!t]
	\begin{footnotesize}
		\caption{The MSE of landmarks, and the number of parameters and the training time of the best two methods (i.e., Yi \emph{et al.}~\cite{yi2020vertebra} and Ours). The evaluation is on 50 images in Consistent split.}
		\label{result_tabel}
		\begin{center}
			\resizebox{1.0\columnwidth}{!}{
			\begin{tabular}{l|c|c|c} 
					\hline
					Methods & MSE & \makecell[c]{Number of \\ Parameters}   & \makecell[c]{Training Time \\  per Epoch } \\
					\hline
					$\mathrm{S^2VR}$~\cite{sun2017direct} & $6.0e{-3}$ & -- & -- \\
					Conv with dense~\cite{Simonyan2015VeryDC} & $7.1e{-2}$ & -- & -- \\
					BoostNet~\cite{wu2017automatic}	& $4.6e{-3}$ & -- & --  \\
					SLSN~\cite{zhang2021automated}	& $3.9e{-3}$ & -- & --  \\
					Yi~\emph{et al.}~\cite{yi2020vertebra}	& $3.2e{-3}$ & 24.21 M & 33.69 s  \\
					Ours & $\bf{7.2e{-4}}$ & 11.01 M & 38.78 s \\
					\hline
				\end{tabular}
			}
		\end{center}
	\end{footnotesize}
\end{table}

\begin{figure}[!t]
	\centering
	\includegraphics[width=0.9\columnwidth]{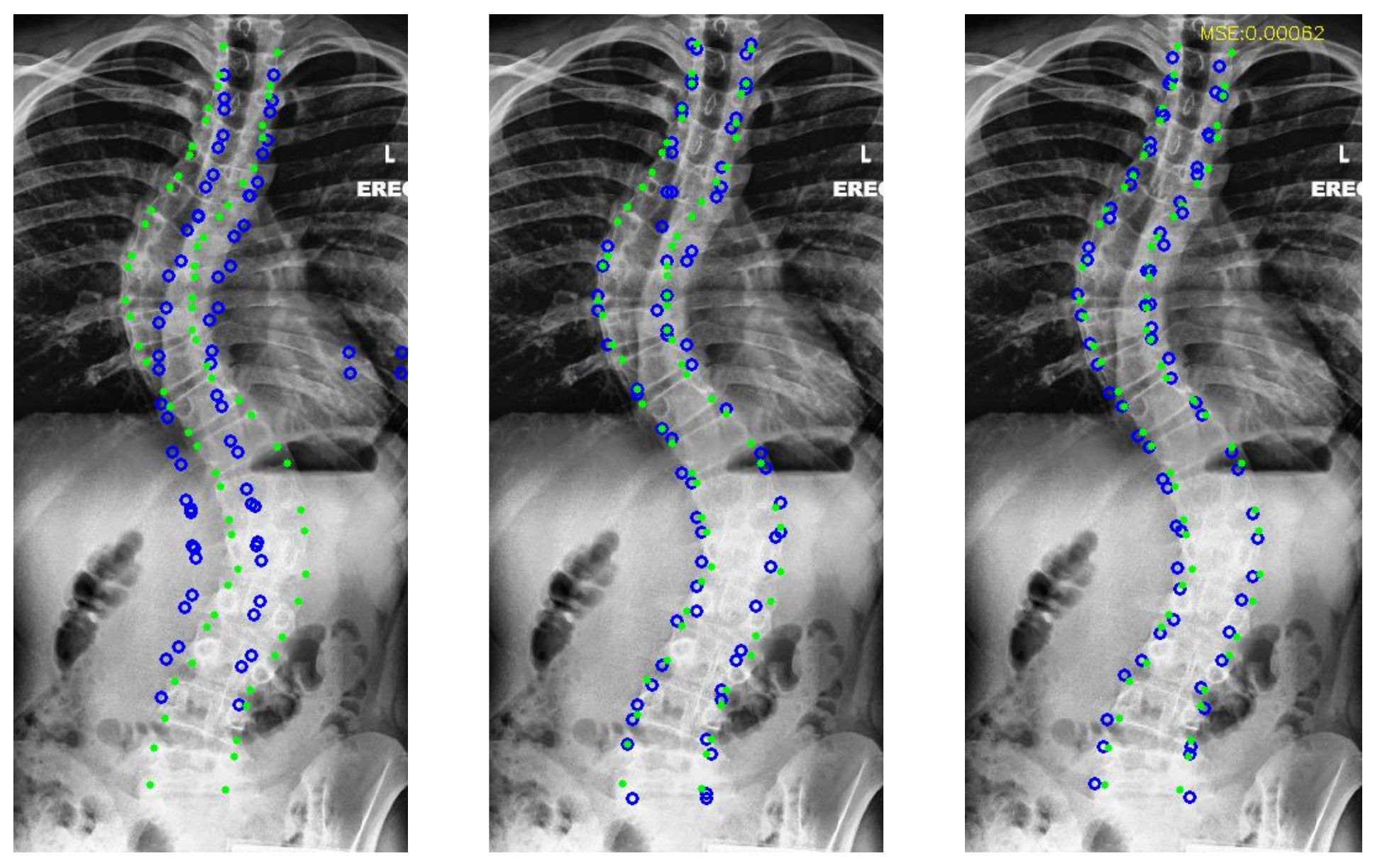}
	\caption{From left to right are the example results of direct regression, heatmap-based regression and ours, respectively. The green dots and blue circles indicate ground truth and prediction, respectively.}
	\label{compare}
\end{figure}

\begin{figure*}[!t]
	\centering
	\includegraphics[width=0.8\textwidth]{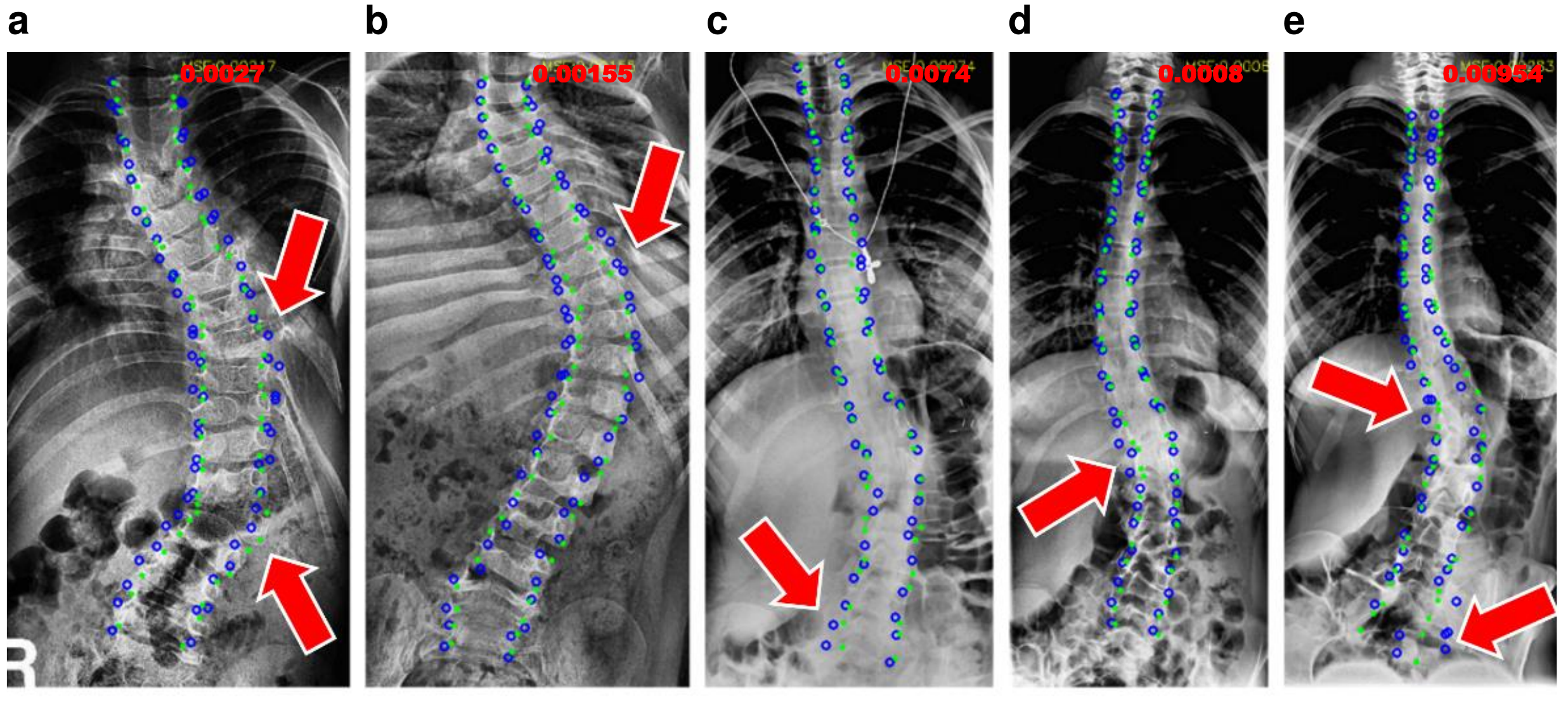}
	\caption{Five failure cases produced by our method. A few is caused by the confusion between vertebrae and ribs (a-b), and the most are caused by the low image contrast (c-e). The green dots and blue circles indicate ground truth and prediction, respectively.}
	\label{fig:6}
\end{figure*}

The second column of Table~\ref{result_tabel} shows the comparison results of different methods.
As can be seen, our method achieves the best performance in terms of MSE ( $7.2e{-4}$), which is near an order of magnitude smaller than~\cite{sun2017direct,wu2017automatic,zhang2021automated,yi2020vertebra}, and two orders of magnitude smaller than~\cite{Simonyan2015VeryDC}.
We believe that the good performance mostly benefits from the multi-stage cascaded framework and the PCA-mined shape constraint.

Concretely, these compared state-of-the-art methods can be divided into two categories: the direct regression methods~\cite{sun2017direct,Simonyan2015VeryDC,wu2017automatic} and the heatmap-based methods~\cite{yi2020vertebra,zhang2021automated}.
Comparing with the direct regression methods, our method uses multi-stage cascaded framework to progressively predict the center and corner points in two steps, decomposing the difficult task into simpler ones.
Comparing with the heatmap-based methods, our method regresses a few eigenvalues derived by PCA instead of the coordinates, better preserving the shape constraint. In comparison, the heatmap-based methods essentially segment those pixels close to the landmarks, thus sharing the common issues with the segmentation techniques, i.e., scattered false positives incurred by other vertebral-like structures and missing bones induced by dilution of superimposed tissues.

Fig.~\ref{compare} displays the results achieved by direct regression~\cite{Simonyan2015VeryDC} (implemented by ourselves for visualization), heatmap-based approach~\cite{yi2020vertebra} using U-Net as backbone~\cite{payer2016regressing} and our proposed multi-stage cascaded CNNs. 
It can be observed that the performance of direct regression is poor, and the heatmap-based approach almost locates the landmarks while suffers missing localizations induced by dilution of superimposed tissues, e.g., tissues in the lung.
It is worth noting that if we manually suppress false positives, and exclude those missed vertebrae from the evaluation, the error of~\cite{yi2020vertebra} dramatically decreases to $8.1e{-4}$ which is comparable to ours. This well demonstrates that the heatmap-based method suffers from false positives and missing localization in the task of vertebral landmark localization.

{{\bf Failures.}}
We also show several failure cases achieved by our method in Fig.~\ref{fig:6}.
We can observe that most landmarks are successfully localized and the overall shape is correct thanks to the PCA-mined shape constraint.
However, some landmarks are still misaligned with GT, which should be caused by the low image contrast (see Figure~\ref{fig:6}-(c), (d), (e)) and the confusion between vertebrae and ribs (see Figure~\ref{fig:6}-(a), (b)).

{{\bf Computational cost.}}
We further compare the model size and the training time between the best two methods (i.e., Yi \emph{et al.}~\cite{yi2020vertebra} and Ours).
Although our method is a multi-stage approaches, each stage just utilizes a light-weight CNN named MobileNet-v2 for feature extraction which was developed for the purpose of portable devices.
In addition, the last few layers of the original MobileNet-v2 were truncated, further decreasing the number of parameters.
Therefore, each stage just contains around 1.83 M parameters, and 6 stages (half for center point localization and half for corner point localization) totally have a parameter size of 11.01 M as shown in Table~\ref{result_tabel}.

The training time is recorded by running out all data (one epoch) with the batch size equal to 2 on the computational platform with a TITAN XP GPU.
Results in Table~\ref{result_tabel} show that, our method outperforms Yi \emph{et al.}~\cite{yi2020vertebra} in terms of accuracy for vertebral landmark localization and meanwhile requires much fewer parameters, i.e., fewer than half of theirs.
However, since our model split the task into two steps and each step contains three stages, the resulting overhead between every two stages causes slightly more training time than~\cite{yi2020vertebra} consequently (around 5 more seconds per epoch).

\subsubsection{Comparison with challenge participants for Cobb angle estimation}

We compare our method with the top-5 participants of the AASCE challenge.
The challenge participants evaluated their methods using the estimation error of Cobb angles while our method predicts the vertebral landmarks.
To perform a fair comparison, we utilize our localized vertebral landmarks to calculate the three Cobb angles.
The Cobb angle calculation is realized by an official tool written in MATLAB provided by the challenge organizer, which can automatically convert the input landmarks to the three Cobb angles, i.e., PT angle, MT angle and TL angle.

\begin{figure*}[!t]
	\centering
	\includegraphics[width=0.8\textwidth]{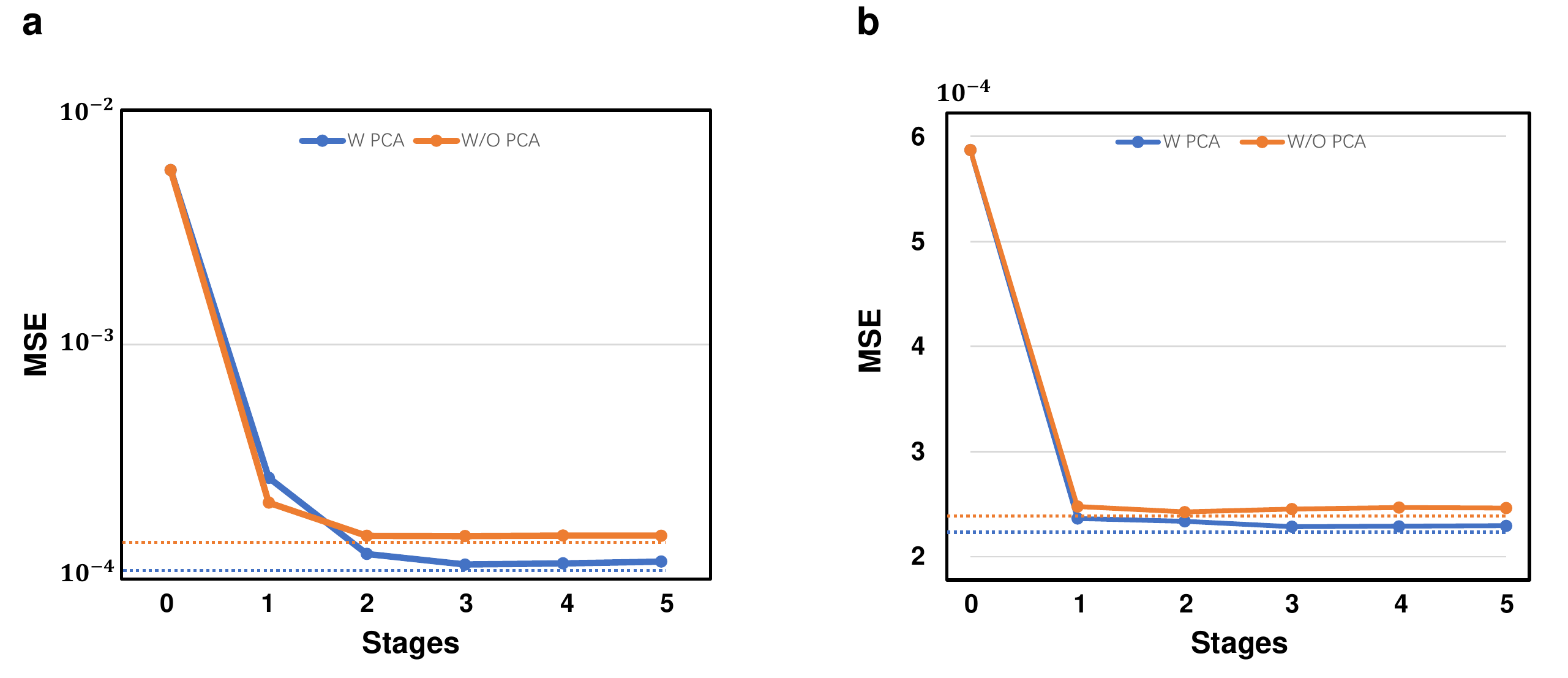}
	\caption{Illustration of the accuracy improvement via PCA based data pre-processing and best choice of stages.(a) represents center points localization sub-task and (b) refers to corner points localization sub-task based on GT centers, respectively. The blue line and orange line show the MSE with PCA and Without PCA respectively.  The horizontal axis stands for the stages.
	}
	\label{Ablation_Figs}
\end{figure*}

We re-train our method based on the Official Split, and evaluate on the 98 challenge-private test images.
For these challenge-private images, the organizer only released X-ray images for the challenge-private test images, but did not give the ground-truth landmarks as well as Cobb angles.
Therefore, we asked several doctors to manually annotate the vertebral landmarks and utilized the official tool to calculate the Cobb angles as the ground-truth ones.

\begin{table}[!t]
	\begin{footnotesize}
		\caption{Comparison results of Cobb angle estimation on the challenge-private test images. Lower SMAPE value indicates better performance.}
		\label{challenge}
		\begin{center}
			\renewcommand\arraystretch{1.2}
			\setlength{\tabcolsep}{2mm}{
				\begin{tabular}{l|c}
					\hline
					Participants & SMAPE \\
					\hline
					Tencent~\cite{lin2019seg4reg}  & 21.71\% \\
					iFLYTEK~\cite{chen2019accurate}  & 22.17\% \\
					XMU~\cite{wang2019spinal}  & 22.18\% \\
					ErasmusMC~\cite{dubost2019automated}  & 22.96\% \\
					XDU~\cite{zhong2019coarse}  & 24.80\% \\
					Ours & $\bf{21.30\%}$ \\
					\hline
				\end{tabular}
			}
		\end{center}
	\end{footnotesize}
\end{table}

Table~\ref{challenge} shows the results between top-5 participants and our method.
Among these participants, three of them (XMU, iFLYTEK and XDU) first localized 68 vertebral landmarks, and then used the official tool to measure the three Cobb angles.
The rest two (Tencent and ErasmusMC) skipped the landmark localization step and directly regressed the Cobb angles based on the whole X-ray image.

The SMAPE achieved by our method is lower than XMU, iFLYTEK and XDU, which implies that the landmarks predicted by our method could be more accurate than theirs.
Moreover, by enjoying more accurate vertebral landmarks, our performance achieves the 1-st place on the Leaderboard\footnote{https://aasce19.grand-challenge.org/evaluation/challenge/leaderboard/}, and surpasses the winner Tencent, which directly regresses the Cobb angles without giving those explainable vertebral landmarks.

\subsubsection{Ablation study}

{{\bf Effectiveness of PCA-mined shape constraint.}}
also validate the effectiveness of PCA-mined shape constraint in the step of center point localization. 
We develop a non-PCA version of our proposed method, where the Coordinate Offset Regression module directly regresses the offsets of coordinates instead of the eigenvalues.
We calculate the smooth L1 distance between the predicted shape offsets and the GT offsets as the loss to optimize the parameters of the shared encoder and the fully connected layer.
Both PCA and non-PCA versions of our method are trained with 5 stages and each stage takes 8 epochs.

\begin{figure}[!t]
	\centering
	\includegraphics[width=1.0\columnwidth]{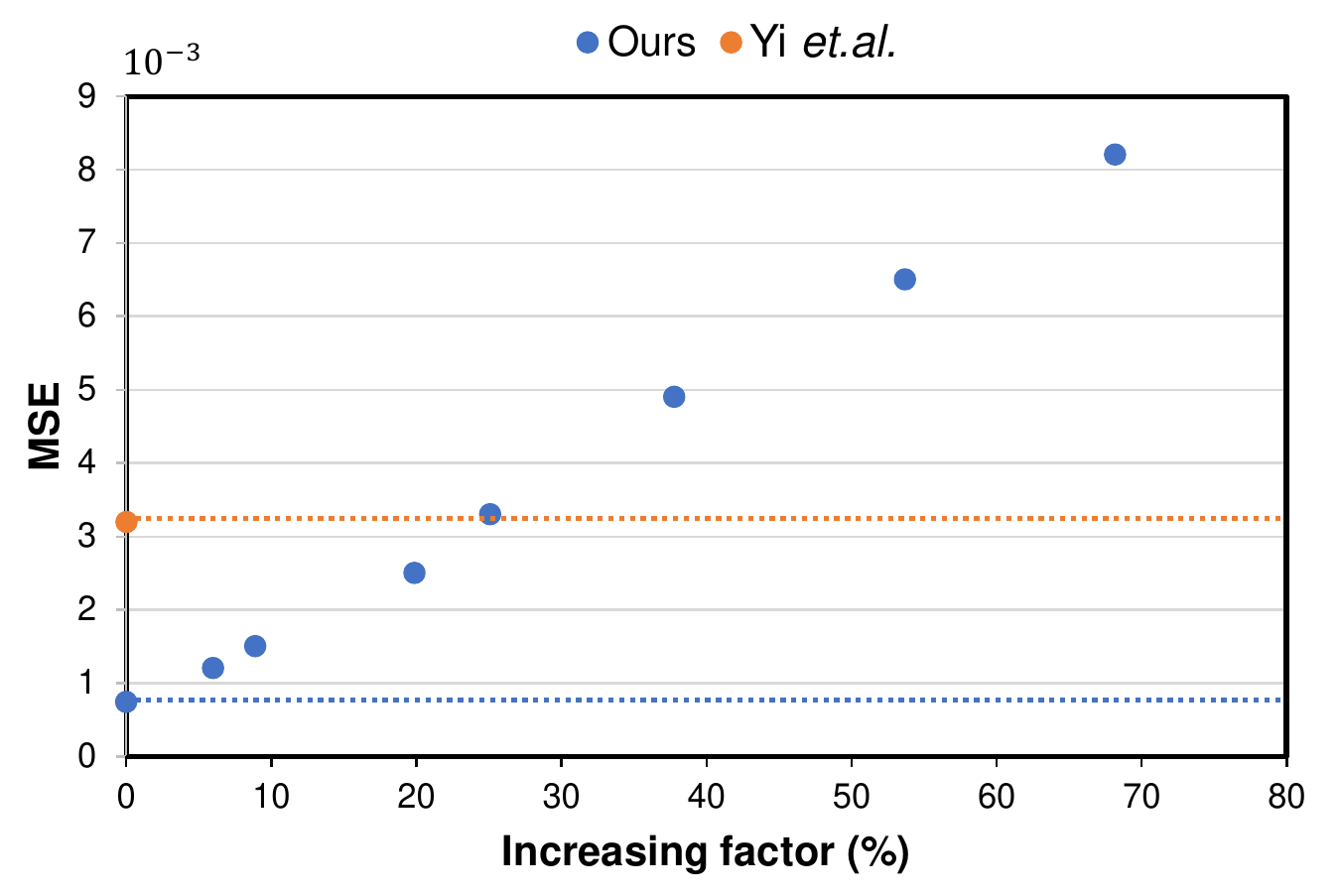}
	\caption{Validation of the sensitivity to the initialization.}
	\label{initialization}
\end{figure}

Fig.~\ref{Ablation_Figs}(a) displays the performance in each stage of the two versions of our method for the task of center points localization.
The horizontal axis stands for the index of stage and the vertical axis stands for the error on the official validation set.
At stage-0, the errors are the same as we utilize the same normalized averaged shape for initialization.
For both versions w/ and w/o PCA, the error decreases firstly and achieves the best performance at stage-3 while slightly increases in the later stages because of the over-fitting problem.
The two dash lines represent the lowest errors achieved by the two versions at the $3$rd stage, respectively, which clearly shows that our proposed PCA-mined shape constraint can boot the performance by a promising margin.
Without PCA, the points move independently under no priori shape constraint, thus misalign with the GT in regions with heavily superimposing tissues or wrongly align with the nearby vertebrae.

We then validate the effectiveness of PCA-mined shape constraint in the step of corner point localization.
Fig.~\ref{Ablation_Figs}(b) illustrates the comparison results between the two versions w/ and w/o PCA, and both are performed on the extracted RoIs at GT center points to eliminate the influence of step-1.
The same conclusions can be observed from Fig.~\ref{Ablation_Figs}(b) that both achieve the best performance at the stage-3, and the method with PCA surpasses that without PCA, which well demonstrates the effectiveness of our proposed PCA-mined shape constraint for the landmark localization.
It is noteworthy that when comparing Fig.~\ref{Ablation_Figs}(a) and (b) at stage-0 the initial error of the center points is much higher than corner points as the relatively higher variance of the spinal shape and fixed structure of the vertebra.

{{\bf Sensitivity to initialization.}}
We further validate the sensitivity of our multi-stage cascaded framework to the initialization.
To this end, we disturb the initial coordinates by adding a random value, which is drawn from a Gaussian distribution $\sim N\left(0,\sigma^2\right)$ to each of the normalized coordinates of the initial shape. 
The standard deviation $\sigma$ controls the severity of distribution. 
By increasing $\sigma$, the MSE between the initialization and the ground-truth increases accordingly, therefore, we can evaluate the sensitivity of our method to different initialization.

As shown in Fig.~\ref{initialization}, the horizontal axis stands for the increasing factor of initial MSE and the vertical axis stands for the error of final estimated corner points on the test images of the Consistent Split. 
The blue dash line indicates that the final MSE is  $7.2e{-4}$ when the initial coordinates with MSE of $1.4e{-2}$ are not disturbed. 
With initial MSE increasing, the final MSE error increases linearly, as shown in blue dots.
It could be seen that our method is relatively robust enough to initialization, and can maintain a superior performance (MSE $< 3.2e{-3}$) to the comparison methods (e.g., orange dash line indicates the result of~\cite{yi2020vertebra}) given distributions less than 24.6\%.

\section{Discussion and Conclusion}
Vertebral landmark localization on X-ray images for scoliosis patients is a crucial step to provide clinically valuable biomarkers for variant downstream applications, e.g., Cobb angle estimation, vertebra identification, etc, but very extremely challenging due to low contrast induced by superimposing tissues and wrongly misalignment with nearby vertebrae.
Existing solutions mainly resort to two approaches, i.e., heatmap-based approach and regression-based approach.
The heatmap-based approach has been widely-used for landmark localization in the field of computer vision and demonstrated to handle objects with large poses, e.g., facial landmarks or human poses.
However, we argue that it is overused since the vertebral column roughly fits a curve with only a slight variation under a clear prior of shape constraint.
On the other hand, the side effect of those heatmap-based approaches has its application on the vertebral landmark localization often suffer from false positives and missing bones.
The regression-based approach can guarantee one-to-one match between the predicted and the GT landmarks, while it often fails to regress 68 vertebral landmarks at once especially for these homogeneous vertebrae.

In this paper, we address the difficult task of vertebral landmark localization by two sequential steps, i.e., first localizing 17 center points of vertebrae and second localizing 68 corner points around the predicted center points in the first step.
By doing so, the difficulty decreases in both step, which can be easily handled by our models.
Especially in the second step, we can have our model focus on every single vertebra, preventing the localization process disturbed by other nearby vertebrae.
Furthermore, in each step, we design a multi-stage cascaded networks with PCA-based shape constraint to further decrease the difficulty, which is inspired by the philosophy of a boosting algorithm named Gradient Boosting Decision Tree (GBDT)~\cite{friedman2003multiple}.
GBDT solves a difficult problem by multiple weak learners in multiple stages, and each leaner learns to fix the unsolved problem remained from the last learner.
Similarly, we trained three CNNs in three stages, and each CNN is only responsible for predicting the offset between the GT and the prediction from the previous stage.
We employed the light-weight MobileNet-V2 as the weak learner, therefore, our method just has a few parameters but achieves a promising performance.

Extensive and comprehensive experiments demonstrated the superior performance of our method in the task of vertebral landmark localization comparing with the existing state-of-the-arts, as well as the task of the downstream Cobb angle estimation comparing with the AASCE challenge participants.
We also demonstrate the effectiveness of our proposed PCA-mined shape constraint for the vertebral landmark localization, which allows us to gradually update an initial shape to the GT shape in several eigenvalue-defined directions, e.g., bending, scaling, etc, instead of  moving landmarks independently and uncontrollably.

The study of failure cases has clearly show that even though there exist a few misaligned landmarks, the rough shape of the vertebral column is still well-preserved.
And those misaligned landmarks are mainly caused by the confusion of ribs and highly superimposed tissues, which verifies the high difficulty of vertebral landmark localization on X-ray images for scoliosis patients.

In the future, we will introduce an end-to-end training strategy into this work, thus make the two steps and cascaded CNNs mutually guide each other, and expect to further boost the performance.
Also, we will extend our approach to other landmark localization tasks especially there are high correlations between the landmarks.

\bibliographystyle{IEEEtran}
\bibliography{ref}

\end{document}